\documentclass[10pt,twocolumn,letterpaper]{article}

\usepackage{cvpr}              

\usepackage{graphicx}
\usepackage{amsmath}
\usepackage{amssymb}
\usepackage{booktabs}
\usepackage{dsfont}
\newcommand{\xcc}[1]{#1}

\usepackage[pagebackref,breaklinks,colorlinks]{hyperref}

\usepackage[capitalize]{cleveref}
\crefname{section}{Sec.}{Secs.}
\Crefname{section}{Section}{Sections}
\Crefname{table}{Table}{Tables}
\crefname{table}{Tab.}{Tabs.}

\def\cvprPaperID{1741} 
\def\confName{CVPR}
\def\confYear{2023}

\begin{document}

\title{Unsupervised Domain Adaption with Pixel-level Discriminator \\ for Image-aware Layout Generation}

\author{Chenchen Xu$^{1,2}$\thanks{Work done during an internship at Alibaba Group}\and 
Min Zhou$^2$\and
Tiezheng Ge$^2$\and 
Yuning Jiang$^2$\and
Weiwei Xu$^1$\thanks{Corresponding author}\\
$^1$State Key Lab of CAD\&CG, Zhejiang University \quad $^2$Alibaba Group\\
{\tt\small
xuchenchen@zju.edu.cn,
\{yunqi.zm, tiezheng.gtz, mengzhu.jyn\}@alibaba-inc.com,
xww@cad.zju.edu.cn}
}
\maketitle

\begin{abstract}
   Layout is essential for graphic design and poster generation. Recently, applying deep learning models to generate layouts has attracted increasing attention. This paper focuses on using the GAN-based model conditioned on image contents to generate advertising poster graphic layouts, which requires an advertising poster layout dataset with paired product images and graphic layouts. However, the paired images and layouts in the existing dataset are collected by inpainting and annotating posters, respectively. There exists a domain gap between inpainted posters (source domain data) and clean product images (target domain data). Therefore, this paper combines unsupervised domain adaption techniques to design a GAN with a novel pixel-level discriminator (PD), called PDA-GAN, to generate graphic layouts according to image contents. The PD is connected to the shallow level feature map and computes the GAN loss for each input-image pixel. Both quantitative and qualitative evaluations demonstrate that PDA-GAN can achieve state-of-the-art performances and generate high-quality image-aware graphic layouts for advertising posters.
\end{abstract}

\section{Introduction}
\label{sec:intro}
    \begin{figure}
    \centering
    \includegraphics[width=8cm]{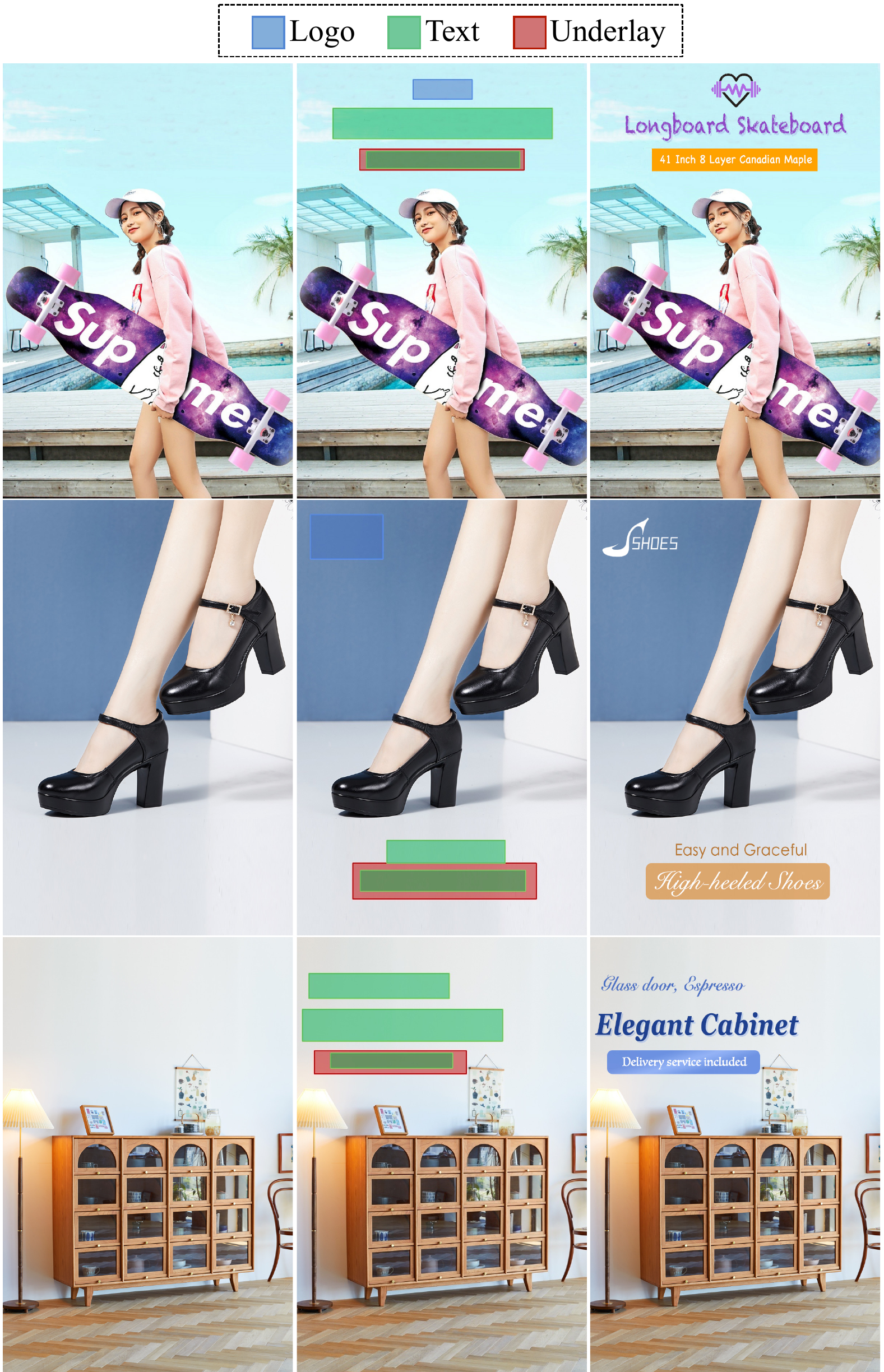}
    \caption{{\bf Examples of image-conditioned advertising posters graphic layouts generation. } Our model generates graphic layouts (middle) with multiple elements conditioned on product images (left).  The designer or even automatic rendering programs can utilize graphic layouts to render advertising posters (right).}
    \label{fig:introduction}
    \end{figure}
  Graphic layout is essential to the design of posters, magazines, comics, and webpages. Recently, generative adversarial network (GAN) has been applied to synthesize graphic layouts through modeling the geometric relationships of different types of 2D elements, for instance, text and logo bounding boxes  \cite{DBLP:conf/nips/GoodfellowPMXWOCB14,DBLP:conf/iclr/LiYHZX19}. Fine-grained controls over the layout generation process can be realized using Conditional GANs, and the conditions might include image contents and the attributes of graphic elements, e.g. category, area, and aspect ratio \cite{DBLP:journals/tvcg/LiY0LWX21,DBLP:conf/ijcai/ZhouXMGJX22}. Especially, image contents play an important role in generating image-aware graphic layouts of posters and magazines \cite{DBLP:journals/tog/ZhengQCL19,DBLP:conf/ijcai/ZhouXMGJX22}.

 This paper focuses on studying the deep-model based image-aware graphic layout method for advertising poster design, where the graphic layout is defined to be a set of elements with their classes and bounding boxes as in~\cite{DBLP:conf/ijcai/ZhouXMGJX22}. As shown in \cref{fig:introduction}, the graphic layout for advertising poster design in our work refers to arranging four classes of elements, such as logos, texts, underlays, and other elements for embellishment, at the appropriate position according to product images. Therefore, its kernel is to model the relationship between the image contents and layout elements~\cite{DBLP:conf/ijcai/ZhouXMGJX22,DBLP:conf/mm/CaoMZLXGJ22} such that the neural network can learn how to produce the aesthetic arrangement of elements around the product image. It can be defined as the direct set prediction problem in~\cite{DBLP:conf/eccv/CarionMSUKZ20}.
 
 Constructing a high-quality layout dataset for the training of image-ware graphic layout methods is labor intensive, since it requires professional stylists to design the arrangement of elements to form the paired product image and layout data items. For the purpose of reducing workload, zhou et.al.~\cite{DBLP:conf/ijcai/ZhouXMGJX22} propose to collect designed poster images to construct a dataset with required paired data. Hence, the graphic elements imposed on the poster image are removed through image inpainting~\cite{DBLP:conf/wacv/SuvorovLMRASKGP22}, and annotated with their geometric arrangements in the posters, which results in state-of-the-art CGL-Dataset with 54,546 paired data items. While CGL-Dataset is substantially beneficial to the training of image-ware networks, there exists a domain gap between product image and its inpainted version. The CGL-GAN in~\cite{DBLP:conf/ijcai/ZhouXMGJX22} tries to narrow this domain gap by utilizing Gaussian blur such that the network can take a clean product image as input for synthesizing a high-quality graphic layout. However, it is possible that the blurred images lose the delicate color and texture details of products, leading to unpleasing occlusion or placement of graphic elements. 
 

 This paper proposes to leverage  unsupervised domain adaption technique to bridge the domain gap between \xcc{clean product images and inpainted images} in CGL-Dataset to significantly improve the quality of generated graphic layouts. Treating the inpainted poster images without graphic elements as \xcc{the} source domain, our method aims to seek for the alignment of the feature space of \xcc{source domain and the feature space of clean product images in the target domain}. To this end, we design a GAN with \xcc{a} pixel-level domain adaption discriminator, abbreviated as PDA-GAN, to achieve more fine-grained control over feature space alignment. It is inspired by PatchGAN~\cite{DBLP:conf/cvpr/IsolaZZE17}, but non-trivially adapts to pixel-level in our task. First, the pixel-level discriminator (PD) designed for domain adaption can avoid the Gaussian blurring step in~\cite{DBLP:conf/ijcai/ZhouXMGJX22}, which is helpful for the network to model the details of the product image. Second, the pixel-level discriminator is connected to the shallow level feature map, since the inpainted area is usually small relative to the whole image and will be difficult to discern at deep levels with large receptive field. Finally, the PD is constructed by three convolutional layers only, and its number of network parameters is less than $2\%$ of the discriminator parameters in CGL-GAN. This design reduces the memory and computational cost of the PD.
 


 
 We collect 120,000 target domain images during the training of PDA-GAN. Experimental results show that PDA-GAN achieves state-of-the-art (SOTA) performance according to composition-relevant metrics. It outperforms CGL-GAN on CGL-dataset and achieves relative improvement over background complexity, occlusion subject degree, and occlusion product degree metrics by $6.21\%$, $17.5\%$, and $14.5\%$ relatively, leading to significantly improved visual quality of synthesized graphic layouts in many cases. 
 In summary, this paper comprises the following contributions:
 \begin{itemize}
     \item We design a GAN with a novel pixel-level discriminator working on shallow level features to bridge the domain gap that exists between training images in CGL-Dataset and clean product images.
     \item Both quantitative and qualitative evaluations demonstrate that PDA-GAN can achieve SOTA performance and is able to generate high-quality image-aware graphic layouts for advertising posters.
 \end{itemize}

 \section{Related Work}
\paragraph{Image-agnostic layout generation.}
Early works~\cite{DBLP:journals/tog/JacobsLSBS03,DBLP:conf/chi/KumarTAK11,DBLP:journals/tog/CaoCL12,DBLP:journals/tvcg/ODonovanAH14} often utilize templates and heuristic rules to generate layouts. LayoutGAN~\cite{DBLP:conf/iclr/LiYHZX19} is the first method to apply generative networks~(in particular GAN) to synthesize layouts and use self-attention to build the element relationship. LayoutVAE~\cite{DBLP:conf/iccv/JyothiDHSM19} and LayoutVTN~\cite{DBLP:conf/cvpr/ArroyoPT21} follow and apply VAE and autoregressive methods. Meanwhile, some conditional methods have been proposed to guide the layout generation process~\cite{DBLP:conf/eccv/LeeJELG0Y20,DBLP:conf/cvpr/YangFYW21,DBLP:conf/mm/KikuchiSOY21,DBLP:journals/tvcg/LiY0LWX21,DBLP:conf/iccv/GuptaLA0MS21}. The constraints are in various forms, such as scene graphs, element attributes, and partial layouts. 
However, in a nutshell, these methods mainly focus on modeling the internal relationship between graphic elements, and rarely consider the relationship between layouts and images.

\begin{figure*}[t]
\centering
\hspace{0cm}\includegraphics[width=17.6cm]{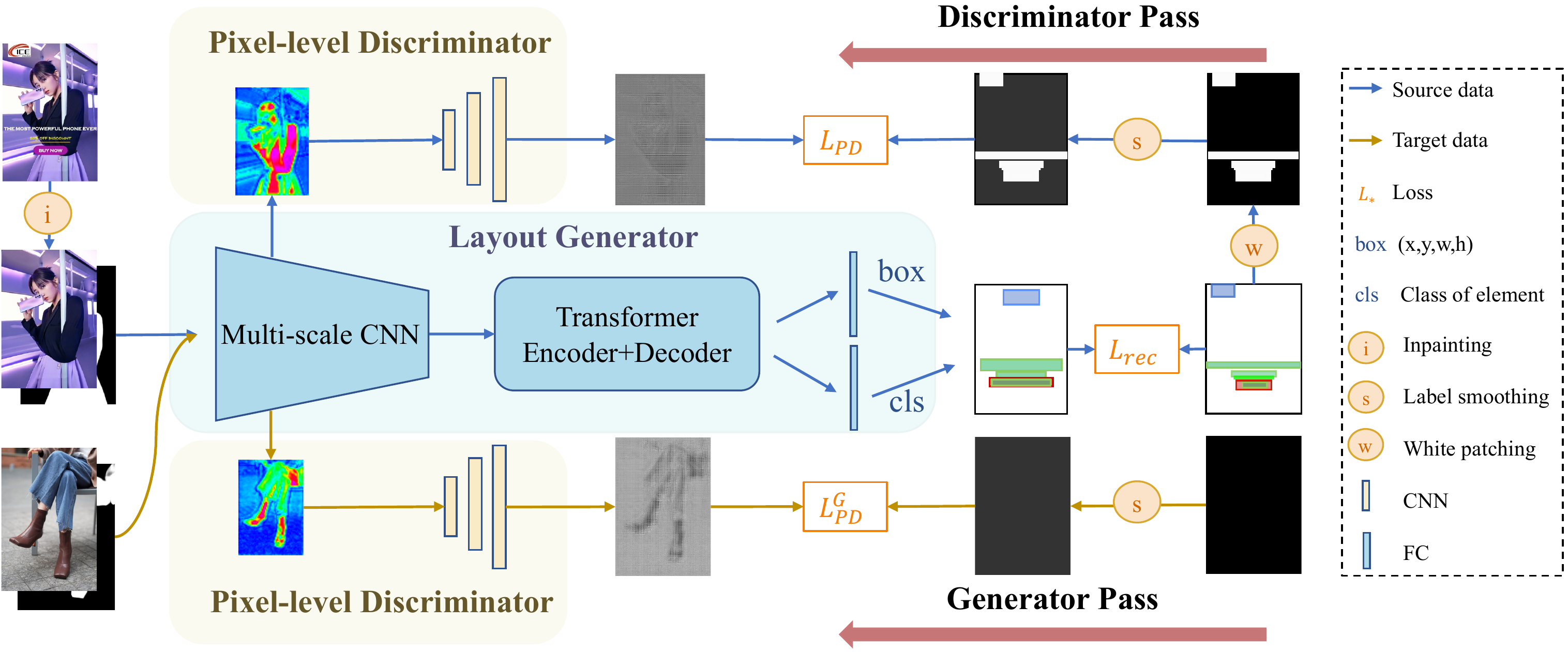}
\caption{{\bf The architecture of our network.} Annotated posters (source domain data) must be inpainted before input to the model. The model has both reconstruction and GAN loss when training with source domain data, while only has a GAN loss is used when training with target domain data. Please refer to Sec. \ref{section.3} for the definition of each loss term: $L_{PD}$, $L_{PD}^G$, and $L_{rec}$. During the discriminator or generator pass, both inpainted and clean images are fed into the discriminator.}
\label{fig:model}
\end{figure*}

\paragraph{Image-aware layout generation.}
In layout generation for magazine pages, ContentGAN~\cite{DBLP:journals/tog/ZhengQCL19} first proposes to model the relationship not only between layout elements but also between layouts and images. However, the high-quality training data is relatively rare, since it requires professional stylists to design layouts for obtaining paired clean images and layouts. ContentGAN uses white patches to mask the graphic elements on magazine pages, and replaces the clean images with the processed pages for training. For the same problem, in poster layout generation, CGL-GAN~\cite{DBLP:conf/ijcai/ZhouXMGJX22} leverages inpainting to erase the graphic elements on posters, and subsequently applies Gaussian blur on the whole poster to eliminate the inpainting artifacts. The blur strategy effectively narrows the domain gap between inpainted images and clean images, but it may damage the delicate color and texture details of images and leads to unpleasing occlusion or element placement. In this paper, we find that a pixel-level discriminator for domain adaption can achieve the same goal and avoid the negative effects of blur.

\paragraph{Unsupervised domain adaptation.}
Unsupervised domain adaptation~\cite{DBLP:journals/corr/abs-2010-03978} aims at aligning the disparity between domains such that a model trained on the source domain
with labels can be generalized into the target domain, which lacks labels. Many related methods~\cite{DBLP:conf/esann/MajumdarN18,DBLP:journals/corr/abs-2205-12923,DBLP:journals/mta/ZhangD21,DBLP:conf/cvpr/Zhang0TL22,DBLP:conf/cvpr/BousmalisSDEK17,DBLP:journals/corr/abs-2010-03978,DBLP:conf/aaai/PeiCLW18,DBLP:journals/tip/RenLZH22} have been applied for object recognition and detection. Among these methods, \cite{DBLP:conf/cvpr/BousmalisSDEK17,DBLP:journals/corr/abs-2010-03978,DBLP:conf/aaai/PeiCLW18,DBLP:journals/tip/RenLZH22} leverage adversarial domain adaptation approach~\cite{DBLP:series/acvpr/GaninUAGLLML17}. A domain discriminator is employed and outputs a probability value indicating the domain of input data. In this way, the generator can extract domain-invariant features and eliminate the semantic or stylistic gap between the two domains. However, it does not work well when applied directly to our problem, since the inpainted area is small compared to the whole image and is difficult to discern at deep levels. Therefore, we design a pixel-level discriminator to effectively solve this.

 \section{Our Model} \label{section.3}
 
 Our model is a generative adversarial network to learn domain-invariant features with the pixel-level discriminator to minimize the cross-domain discrepancy. As shown in \cref{fig:model}, our network mainly has two sub-networks: the layout generator network that takes the image and its saliency map as the input to generate graphic layout and the convolutional neural network for pixel-level discriminator.
 
 In this section, we will describe the details of our network architecture and the training loss functions for the pixel-level discriminator and the layout generator network respectively.

 \subsection{Network Architecture}
 The architecture of the layout generator network is the same with the generator network in~\cite{DBLP:conf/ijcai/ZhouXMGJX22}, but  the user-constraints are ignored. Its design follows the principle of DETR~\cite{DBLP:conf/eccv/CarionMSUKZ20}, which has three modules: a multi-scale convolutional neural network (CNN) used to extract image features  \cite{DBLP:conf/cvpr/HeZRS16,DBLP:conf/cvpr/LinDGHHB17}, a transformer encoder-decoder that accepts layout element queries as input to model the relationship among layout elements and the product image \cite{DBLP:conf/nips/VaswaniSPUJGKP17}, and two fully connected layers to predict the element class and its bounding box using the element feature output by the transformer decoder.  
  
 Our pixel-level discriminator network consists of three transposed convolutional layers with filter size $3\times 3$ and stride $2$. Its input is the feature map from the first residual block in multi-scale CNN. The transposed convolutional layers can up-sample the feature map, and we also allow to resize the final result to exactly match the dimension of the input image to facilitate the computation of discriminator training loss, which will be elaborated in the next section.

 
 
 
 \begin{table*}[t!]
    \centering
    \setlength{\tabcolsep}{2.18mm}{
    \scalebox{0.98}{
    \begin{tabular}{lccc|cccc}
    \toprule
         Model    &$R_{com}\downarrow$ &$R_{shm}\downarrow$ &$R_{sub}\downarrow$ &$R_{ove}\downarrow$ &$R_{und}\uparrow$ &$R_{ali}\downarrow$ &$R_{occ}\uparrow$\\
    \midrule
         ContentGAN~\cite{DBLP:journals/tog/ZhengQCL19}      &45.59 &17.08 &1.143 &0.0397 &0.8626 &\bf{0.0071} &93.4\\
         CGL-GAN~\cite{DBLP:conf/ijcai/ZhouXMGJX22} &\underline{35.77} &\underline{15.47} &\underline{0.805} &\bf{0.0233} &\underline{0.9359} &\underline{0.0098} &\underline{99.6}\\
         PDA-GAN(Ours)   &\bf{33.55} &\bf{12.77} &\bf{0.688} &\underline{0.0290} &\bf{0.9481} &0.0105   &\bf{99.7}\\
    \bottomrule
    \end{tabular}
    }
    }
    \caption{{\bf Comparison with content-aware methods.} Bold and underlined numbers denote the best and second best respectively.}
    \label{tab:content-aware}
\end{table*}
 
\subsection{Pixel-level Discriminator Training} 
The design of pixel-level discriminator is based on the observation that the domain gap between inpainted images and clean product images mainly exists at pixels synthesized by inpainting process. Therefore, during the discriminator or generator pass in \cref{fig:model}, both inpainted and clean images are fed into the discriminator. When updating the discriminator, we encourage the discriminator to detect the inpainted pixels for inpainted images in the source domain. In contrast, when updating the generator, we leverage the pixel-level discriminator to encourage the generator to output shallow feature maps that can fool the discriminator, which means that, even for the feature map computed for the inpainted images, the discriminator's ability to detect inpainted pixels should be weakened fast. In this way, when the training converges, the feature space of source and target domain images should be aligned. 

To calculate the loss $L_{PD}$ for each input-image pixel, we utilized the white-patch map to distinguish whether the input-image pixel is inpainted, where the pixel in white patch map is set to 1 if the corresponding pixel in the input image is processed by the inpainting, otherwise 0. Correspondingly, the pixel values of white-patch map for the clean images in target domain are all 0. 

When updating discriminator in the GAN training, the pixel-level discriminator takes shallow level feature maps as input and outputs a map with one channel whose dimension is consistent with the input image. The loss $L_{PD}$ used to train the discriminator is a mean absolute error (MAE) loss or L1 norm between the white-patch map of input images and the output map. We can get  as:
 \begin{equation}
    \begin{aligned}
    {L}_{PD} = 
    \frac{1}{N_{p}} \sum^{N_{p}}_{i=1} (
    {\left\vert {{\mathbf{P}}^{s,w}_i - {\mathbf{P}}^{s,o}_i} \right\vert} * \alpha \\ + {\left\vert {{\mathbf{P}}^{t,w}_i - {\mathbf{P}}^{t,o}_i} \right\vert} * \beta),
    \end{aligned}
\end{equation}
where the $N_{p}$ means the number of white-patch map pixels, and $\mathbf{P}_i$ indicates the predicted or ground-truth map for $i_{\text{th}}$ image. The superscript of $\mathbf{P}_i$ indicates it is from source by $s$  or target by $t$, from prediction by $o$ or ground truth by $w$. The two coefficients, $\alpha$ and $\beta$, are used to balance between the source and target domain white-patch map. Since the area of the inpainted pixels in the white-patch map are usually small, we set the value of $\alpha$ to $2$ and $\beta$ to $1$.

We utilize one-side label smoothing \cite{DBLP:conf/cvpr/SzegedyVISW16,DBLP:journals/corr/Goodfellow17} to improve the generalization ability of the trained model. Since the inpainted areas occupy a small proportion of the input image, we only do label smoothing for pixels not in the inpainted area (those pixels with value 0 in the white patch map), denoted as one-target label smoothing in our experiments. Precisely, we only set $0$ to $0.2$ in the ground truth white patch map.


\subsection{Layout Generator Network Training}
When updating the generator network in the GAN training, we expect to fool the updated discriminator in the detection of inpainted pixels. Therefore, the loss $L_{PD}$ is modified to penalize the generator network if the discriminator outputs pixels with value $1$. Thus, we have:
  \begin{equation}
    \begin{aligned}
    {L}^{G}_{PD} =
    \frac{1}{N_{p}} \sum^{N_{p}}_{i=1} (
    {\left\vert {\hat{\mathbf{P}}^{s}_i - {\mathbf{P}}^{s,o}_i} \right\vert} * \alpha \\ + {\left\vert {{\mathbf{P}}^{t,w}_i - {\mathbf{P}}^{t,o}_i} \right\vert} * \beta),
    \end{aligned}
\end{equation}
where the values of pixels in $\hat{\mathbf{P}}^{s}_i$ are all set to \xcc{$0.2$}. 
The training loss for the layout generator network is as follows:
\begin{equation} \label{eq3}
    {L}_{G} = L_{rec} + \gamma * L^{G}_{PD},
\end{equation}
where the value of the weight coefficient $\gamma$ is set to 6, and the $L_{rec}$ is the reconstruction loss to penalize the deviation between the graphic layout generated by the network and the annotated ground-truth layout for the inpainted images in the source domain.  We calculate the reconstruction loss $L_{rec}$ as the direct set prediction loss in \cite{DBLP:conf/eccv/CarionMSUKZ20}.

\section{Experiments}
In this section, we mainly compare our model with SOTA layout generation methods and its ablation studies. More additional experimental analyses and designed advertising posters using generated layouts can be found in the supplementary materials.

  \begin{figure*}[!t]
    \centering
    \hspace{0cm}\includegraphics[width=16.6cm]{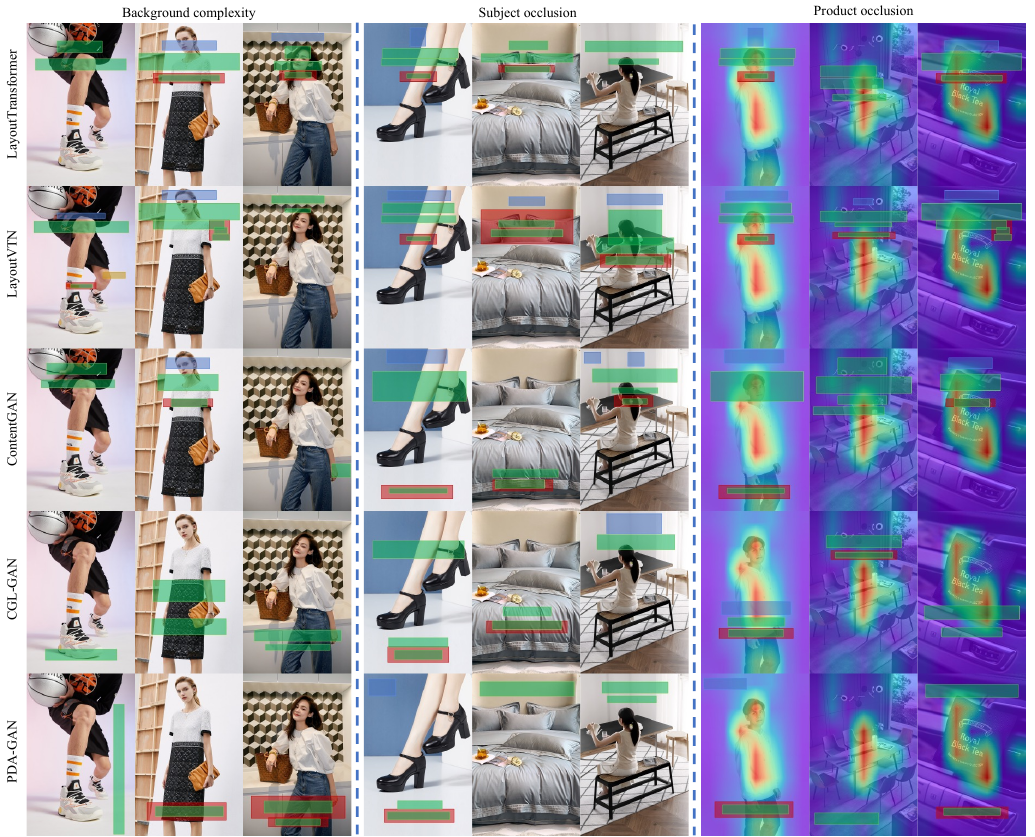}
    \caption{{\bf Qualitative evaluation for different models.} Layouts in a column are conditioned with the same image. And those in a row are from the same model. This figure qualitatively compares and analyzes different models from three aspects: text element background complexity, overlapping subject and overlapping product attention map.}
    \label{fig:3}
\end{figure*}

\begin{table}[!t]
    \centering
    \setlength{\tabcolsep}{0.6mm}{
    \scalebox{1.0}{
    \begin{tabular}{lccc|cccc}
    \toprule
         Model   &$R_{com}\downarrow$ &$R_{shm}\downarrow$ &$R_{sub}\downarrow$ &$R_{ove}\downarrow$ &$R_{und}\uparrow$ &$R_{ali}\downarrow$\\
    \midrule
         LT     &40.92 &21.08 &1.310 &0.0156 &0.9516 &0.0049\\
         VTN      &41.77 &22.21 &1.323 &\bf{0.0130} &\bf{0.9698} &\bf{0.0047}\\
         Ours     &\bf{33.55} &\bf{12.77} &\bf{0.688} &0.0290 &0.9481 &0.0105\\
    \bottomrule
    \end{tabular}
    }
    }
    \caption{{\bf Comparison with content-agnostic methods.} $LT$ and $VTN$ represent LayoutTransformer and LayoutVTN, repectively.}
    \label{tab:content-unaware}
\end{table}

  \begin{figure*}[!h]
    \centering
    \hspace{0cm}\includegraphics[width=15.6cm]{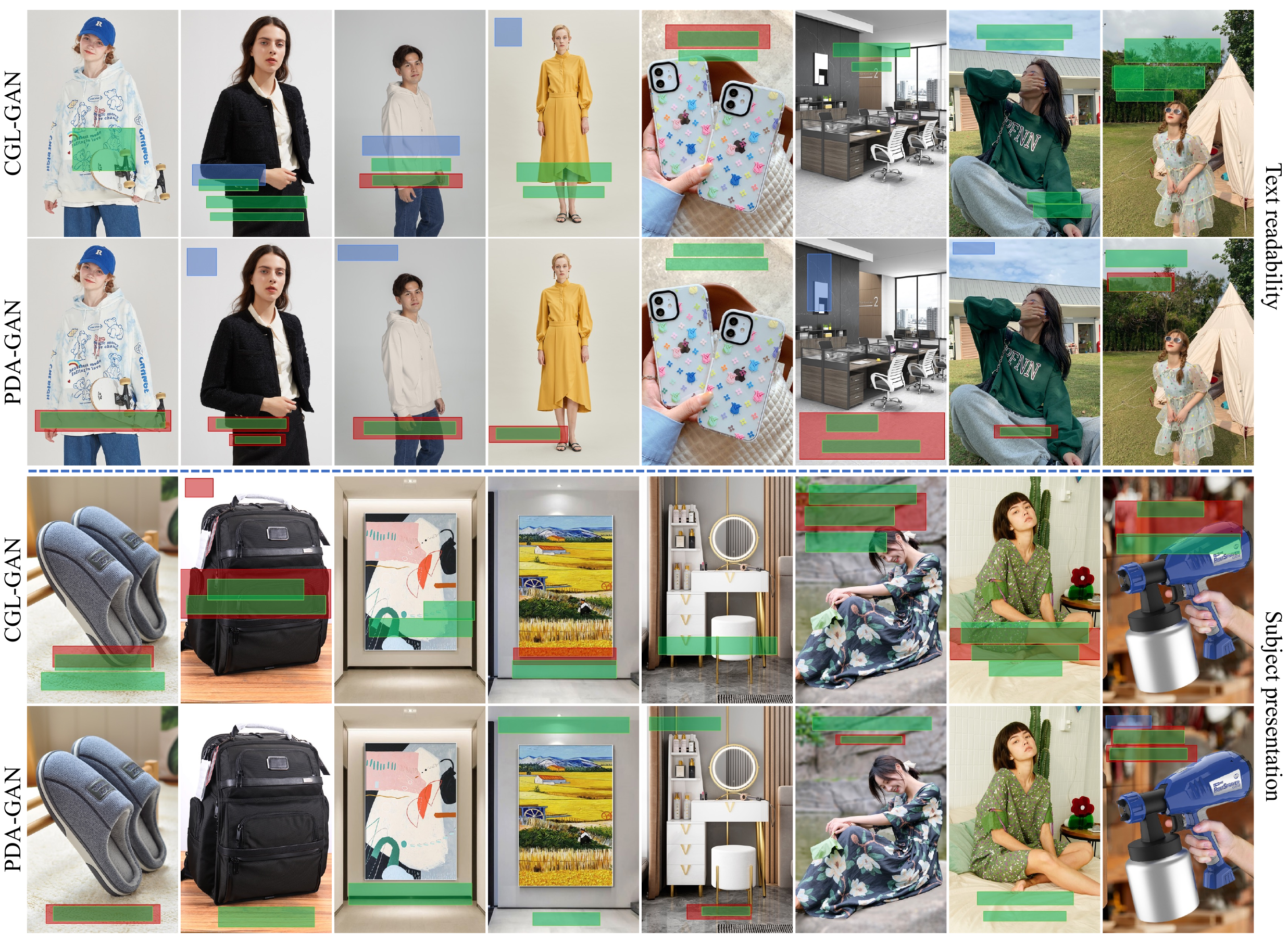}
    \caption{More qualitative comparisons with CGL-GAN.}
    \label{fig:CGL-PDA}
\end{figure*}

\begin{table*}[!t]
    \centering
    \setlength{\tabcolsep}{2.18mm}{
    \scalebox{0.88}{
    \begin{tabular}{lccc|ccc|cccc}
    \toprule
         Model &Data \uppercase\expandafter{\romannumeral1} &Data \uppercase\expandafter{\romannumeral2} &Gaussian Blur  &$R_{com}\downarrow$ &$R_{shm}\downarrow$ &$R_{sub}\downarrow$ &$R_{ove}\downarrow$ &$R_{und}\uparrow$ &$R_{ali}\downarrow$ &$R_{occ}\uparrow$\\
    \midrule
         CGL-GAN &\checkmark &\quad &\quad      &33.85 &13.88 &0.766 &0.0299 &0.9351 &0.0139 &\bf{99.7}\\
         CGL-GAN &\checkmark &\quad &\checkmark      &- &- &- &2.5826 &- &- &-\\

         CGL-GAN &\quad &\checkmark &\checkmark  &35.77 &15.47 &0.805 &\bf{0.0233} &0.9359 &\bf{0.0098} &99.6\\
         PDA-GAN~(Ours) &\checkmark &\quad &\quad   &\bf{33.55} &\bf{12.77} &\bf{0.688} &0.0290 &\bf{0.9481} &0.0105 &\bf{99.7}\\
    \bottomrule
    \end{tabular}
    }
    }
    \caption{{\bf Comprehensive comparison between CGL-GAN and PDA-GAN~(Ours).} Data \uppercase\expandafter{\romannumeral1} and Data \uppercase\expandafter{\romannumeral2} contain 8,000 and 54,546 source domain samples, respectively. \checkmark indicates the experiment configuration. The symbol "-" indicates that the model cannot complete the layout generation task since the generated element bounding boxes overlap with each other severely.
    }
    \label{tab:CGL-GAN}
\end{table*}

\subsection{Implementation Details}
We implement our PDA-GAN in PyTorch and use Adam optimizer for training the model. 
\xcc{The initial learning rates are $10^{-5}$ for the generator backbone and $10^{-4}$ for the remaining part of this model.} The model is trained for 300 epochs with a batch size of 128, and all learning rates are reduced by a factor of 10 after 200 epochs. 
To make the fair experimental comparisons, 
\xcc{we follow} CGL-GAN~\cite{DBLP:conf/ijcai/ZhouXMGJX22} to resize the inpainted posters and product images to $240 \times 350$ as inputs of our PDA-GAN. The total training time is about 8 hours using 16 NVIDIA V100 GPUs.

\xcc{We observe that, during training, the network is prone to bias towards source domain data. It might be due to the additional reconstruction loss for the source domain to supervise the generator of the model. Therefore, to balance the influence of the two domains, 8000 samples are randomly selected from CGL-Dataset as the source domain data. In each epoch, the 8000 source domain samples are processed, and another 8000 samples of the target domain images are randomly selected. We refer to this choice of training data as Data I. If all the CGL-Dataset training images are used for a comparison, we refer to it as Data II. In the following, if not clearly mentioned, our model is trained with Data I.}

 \subsection{Metrics} 
 \xcc{For} quantitative evaluations, we follow \cite{DBLP:conf/ijcai/ZhouXMGJX22} to divide layout metrics into \xcc{the} composition-relevant metrics and \xcc{the} graphic metrics. \xcc{The} composition-relevant metrics include $R_{com}$, and $R_{shm}$, $R_{sub}$, which measure background complexity, occlusion subject degree, and occlusion product degree respectively; \xcc{while the} graphic metrics include $R_{ove}$, $R_{und}$, and $R_{ali}$, which measure layout overlap, underlay overlap, and layout alignment degree respectively. 
 When the $R_{ove}$ value of a model exceeds 0.05, the generated element bounding boxes will overlap each other severely, resulting in useless layouts. This means that the high value of $R_{ove}$ indicates a failure in the layout generation for most images.
 \xcc{Moreover, we use metric $R_{occ}$ to represent} the ratio of non-empty layouts predicted by models. We will use all the above metrics to compare each group's experiments to verify \xcc{the effectiveness of our model}.
 \xcc{The formal definitions of these metrics and corresponding examples of graphic layouts are shown in the supplementary material.}

 \subsection{Comparison with State-of-the-art Methods}
 \noindent\textbf{Layout generation with image contents.} 
 \xcc{We first conduct experiments to} compare our method with ContentGAN and CGL-GAN that can generate image-aware layouts. The quantitative results can be seen from \cref{tab:content-aware}. Our model achieves the best results in most metrics, especially in the composition-relevant metrics since PDA-GAN preserves the image color and texture details. For instance,
 \xcc{our PDA-GAN outperforms contentGAN and CGL-GAN by $26.4\%$ and $6.21\%$ respectively, with regard to background complexity $R_{com}$.}
 As shown in the first column in \cref{fig:3}, compared with \xcc{these by} contentGAN and CGL-GAN, bounding boxes of text element generated by PDA-GAN are more likely to appear in simple background areas, which improves the readability of the text information. As shown in the second and third columns, when the background of the text element is complex, PDA-GAN will generate an underlay bounding box to replace the complex background to \xcc{enhance} the readability of text information.

 \xcc{Comparing contentGAN and CGL-GAN, our PDA-GAN reduces the occlusion subject degree $R_{shm}$ by $25.2\%$ and $17.5\%$ respectively.} From the middle three columns of \cref{fig:3}, 
 \xcc{for contentGAN or CGL-GAN, the presentation of the subject content information are largely affected since the generated layout bounding boxes would inevitably occlude subjects.}
 In particular, it should be noted that when the layout bounding box occludes the critical regions of the subject, such as the human head or face, the visual effect of the poster will be unpleasing, 
 \xcc{taking the image in row-3-column-6 as an example.}
 In contrast, layout bounding boxes generated by PDA-GAN avoid subject regions nicely, \xcc{thus the} generated posters better express the information of subjects and layout elements.
 
 Meanwhile, the occlusion product degree $R_{sub}$ of PDA-GAN performance \xcc{surpass} contentGAN and CGL-GAN by $39.8\%$ and $14.5\%$ respectively. The three rightmost columns in \cref{fig:3} are the heat maps of the attention of each pixel to the product in the image. \xcc{We get attention maps of product images (queried by their category tags extracted on product pages) by CLIP \cite{DBLP:conf/icml/RadfordKHRGASAM21,DBLP:conf/iccv/CheferGW21}.} Compared with contentGAN and CGL-GAN, PDA-GAN generates layout bounding boxes on the region with lower thermal values to avoid occluding products. For example, in the seventh column, the layout bounding box generated by PDA-GAN effectively avoids the region with high thermal values of the product, which makes the hoodie information of the product can be fully displayed.
 The above quantitative and qualitative comparisons of models demonstrate that PDA-GAN improves the relationship modeling between image contents and graphic layouts.

 \noindent\textbf{Layout generation without image contents.} We also compare with image-agnostic methods of LayoutTransformer~\cite{DBLP:conf/iccv/GuptaLA0MS21} and LayoutVTN~\cite{DBLP:conf/cvpr/ArroyoPT21}. As shown in \cref{tab:content-unaware}, these image-agnostic methods perform pretty well on graphic metrics. \xcc{However, in term of composition-relevant metrics, our model is much better. In detail, our PDA-GAN beyonds LayoutTransformer and LayoutVTN by $18.0\%$ and $19.7\%$ respectively with regard to $R_{com}$. That is because these image-agnostic methods only care about the relationship between elements while do not account for image contents.} 
These image-agnostic methods \xcc{are prone to} generate bounding boxes of text elements in the area with complex backgrounds(as shown in the first two rows and \xcc{the leftmost} three columns of \cref{fig:3}), which will reduce the readability of the text information.  \xcc{Furthermore, }compared with LayoutTransformer and LayoutVTN, $R_{shm}$ of PDA-GAN is reduced by $39.4\%$ and $42.5\%$, and $R_{sub}$ is reduced by $47.5\%$ and $48.0\%$. The rightmost six columns in \cref{fig:3} show image-agnostic methods generate layout bounding boxes that randomly occlusion on the subject and product areas. These bounding boxes of layout elements will diminish the content and information presentation of the subject and product.

 \noindent\textbf{More comparisons with CGL-GAN.} As shown in the first and the last row of \cref{tab:CGL-GAN}, PDA-GAN outperforms CGL-GAN in all metrics with the same configuration. 
 \xcc{PDA-GAN differs from CGL-GAN that it uses PD to replace the discriminator in CGL-GAN.}
 The number of the PD parameters is 332,545, less than $2\%$ of the discriminator (22,575,841) in CGL-GAN, which significantly reduces the memory and computation cost of PDA-GAN. 
 The second row of \cref{tab:CGL-GAN} shows that the model training on Data \uppercase\expandafter{\romannumeral1} with Gaussian blur performs poorly on $R_{ove}$, which causes most bounding boxes to overlap each other. 
 \xcc{Intuitively, }Gaussian blur can narrow the domain gap, but it will also cause the image color and texture details lost. From the last two rows of \cref{tab:CGL-GAN}, PDA-GAN without Gaussian blur is far better than CGL-GAN with Gaussian blur on composition-relevant metrics. 
 \xcc{As illustrated in the the sixth and eighth columns of the first two rows in \cref{fig:CGL-PDA}, compared with CGL-GAN, }
 PDA-GAN generates text bounding boxes with the simpler background. 
 \xcc{It is interesting to observe from the first two rows of \cref{fig:CGL-PDA} that when PDA-GAN generates box among complexity background, it tends to additionally generate an underlay bounding box which covers the complex background to ensure the readability of the text information.}
 The last two \xcc{rows} show that layouts generated by PDA-GAN can effectively avoid the subject area, \xcc{and} then \xcc{can generate} posters better express the information of subjects and layout elements.
 
  Both above quantitative and qualitative evaluations demonstrate that PDA-GAN can capture the subtle interaction between image contents and graphic layouts and achieve the SOTA performance. Refer to the supplementary for more details.
 

 \subsection{Ablations}

 \noindent\textbf{Effects of pixel-level Discriminator.} We first compare our PD with a global discriminator that only predicts one real or fake probability as in classical GAN. The abbreviation DA in \cref{tab:PDA-DA} indicates the global discriminator strategy. When the weight of DA loss ($\gamma$ in \cref{eq3}) is more than 0.01, the model cannot complete the layout generation task, indicated by the symbol $-$, since the $R_{ove}$ value is too high. From the statistics in \cref{tab:PDA-DA}, our PD outperforms DA on all metrics. 
 
Second, we compare the PD with the strategy in PatchGAN~\cite{DBLP:conf/cvpr/IsolaZZE17}. The scores of quantitative metrics listed in \cref{tab:PDA-PatchDA} also verify the advantage of PD. The patch size in this table means the dimension of the output map, which will be compared with correspondingly resized ground-truth white patch map during training. These experiments show that, since the discrepancy by inpainting exists between pixels, the model might be required to eliminate the domain gap at the pixel level. Moreover, the pixel-level strategy can be considered as the most fine-grained patch level strategy. 


 \begin{table}[!t]
    \centering
    \setlength{\tabcolsep}{0.58mm}{
    \scalebox{0.96}{
    \begin{tabular}{l|ccc|cccc}
    \toprule
         Model-$W$  &$R_{com}\downarrow$ &$R_{shm}\downarrow$ &$R_{sub}\downarrow$ &$R_{ove}\downarrow$ &$R_{und}\uparrow$ &$R_{ali}\downarrow$\\
    \midrule
         DA-6.0      &- &- &- &9.0000 &- &-\\
         DA-1.0      &- &- &- &8.9995 &- &-\\
         DA-0.01      &- &- &- &4.7764 &- &-\\
         DA-0.001      &34.41 &13.78 &0.749 &0.0327 &0.9299 &0.0110\\
         DA-0.0001     &34.77 &14.62 &0.777 &0.0345 &0.9234 &0.0122\\
         DA-0.0     &34.07 &15.13 &0.800  &0.0350 &0.9259 &0.0108\\

         PDA-6.0   &\bf{33.55} &\bf{12.77} &\bf{0.688} &\bf{0.0290} &\bf{0.9481} &\bf{0.0105}\\
    \bottomrule
    \end{tabular}
    }
    }
    \caption{{\bf Ablation study with discriminator level.} DA indicates the global discriminator strategy in classical GAN-based methods, which outputs one probability value for real or fake for an image. The PDA output is a pixel-level map, and its dimension is same with the input image.  $W$ refers to the weight of DA (or PDA) module loss in the training process. Please refer to ~\cref{tab:CGL-GAN} for the explanation of the symbol "-".}
    \label{tab:PDA-DA}
\end{table}

\begin{table}[!t]
    \centering
    \setlength{\tabcolsep}{0.58mm}{
    \scalebox{0.96}{
    \begin{tabular}{l|ccc|cccc}
    \toprule
         Patch size  &$R_{com}\downarrow$ &$R_{shm}\downarrow$ &$R_{sub}\downarrow$ &$R_{ove}\downarrow$ &$R_{und}\uparrow$ &$R_{ali}\downarrow$\\
    \midrule
         12*8      &- &- &- &0.9288 &- &-\\
         24*16      &33.67 &16.00 &0.844 &0.0438 &0.9407 &\bf{0.0075}\\
         44*30      &34.03 &13.02 &0.752 &\bf{0.0284} &0.9377 &0.0119\\
         88*60      &\bf{32.65} &13.35 &0.735 &0.0325 &0.9173 &0.0094\\

         350*240   &33.55 &\bf{12.77} &\bf{0.688} &0.0290 &\bf{0.9481} &0.0105\\
    \bottomrule
    \end{tabular}
    }
    }
    \caption{{\bf Quantitative ablation study PatchGAN-based methods.} Patch size means the size of the map output by the discriminator. The input image height and width are 320 and 240 respectively. We train these models with $\gamma$ in \cref{eq3} equal to 6. Please refer to ~\cref{tab:CGL-GAN} for the explanation of the symbol "-".}
    \label{tab:PDA-PatchDA}
\end{table}

 
 \noindent\textbf{Effects of PD with different level feature maps.} In our model, PD is connected to the shallow level feature maps of the first residual block. We now investigate how the PD works if the deep level feature, i.e. the feature from fourth residual block, and the fused feature in multi-scale CNN~\cite{DBLP:conf/ijcai/ZhouXMGJX22}, i.e. the fusion of first to fourth residual block feature map, are used in the PD. As shown in \cref{tab:PDA_layer}, discriminating with shallow feature map in PDA-GAN can achieve better results in both composition-relevant and graphic metrics on average. Again, this experiment verifies the advantage of our design of PD. Intuitively, bridging the domain gap at early stage of the network might be 
 beneficial to the subsequent model processing.

 \noindent\textbf{Effects of label smoothing.}  \cref{tab:label_smoothing} shows that the model with one-target label smoothing performs better in all metrics than without label smoothing. In addition, the effects of two-side  or one-source label smoothing are not as good as one-target label smoothing on average. For the ground truth map input to the discriminator, the two-side label smoothing means we set 0 to 0.2 and 1 to 0.8, and one-source label smoothing means we only set 1 to 0.8.
 
\begin{table}[!t]
    \centering
    \setlength{\tabcolsep}{0.54mm}{
    \scalebox{0.98}{
    \begin{tabular}{lccc|ccc}
    \toprule
         Feature map   &$R_{com}\downarrow$ &$R_{shm}\downarrow$ &$R_{sub}\downarrow$ &$R_{ove}\downarrow$ &$R_{und}\uparrow$ &$R_{ali}\downarrow$\\
    \midrule
         deep level      &34.22 &13.97 &0.770 &0.0396 &0.9366 &0.0118\\
         fusion      &35.36 &14.54 &0.817 &0.0310 &\bf{0.9513} &0.0117\\
         shallow level     &\bf{33.55} &\bf{12.77} &\bf{0.688} &\bf{0.0290} &0.9481 &\bf{0.0105}\\
    \bottomrule
    \end{tabular}}
    }
    \caption{{\bf Quantitative ablation study on different level feature maps for pixel-level discriminator.}}
    \label{tab:PDA_layer}
\end{table}

 \begin{table}[!t]
    \centering
    \setlength{\tabcolsep}{0.6mm}{
    \scalebox{0.98}{
    \begin{tabular}{lccc|ccc}
    \toprule
         smoothing   &$R_{com}\downarrow$ &$R_{shm}\downarrow$ &$R_{sub}\downarrow$ &$R_{ove}\downarrow$ &$R_{und}\uparrow$ &$R_{ali}\downarrow$\\
    \midrule
         Without      &33.61 &14.04 &0.718 &0.0346 &0.9188 &0.0106\\
         two-side      &33.66 &14.67 &0.794 &0.0334 &0.9297 &0.0098\\
         one-source    &\bf{32.20} &15.23 &0.799 &0.0431 &0.9234 &\bf{0.0085}\\
         one-target     &33.55 &\bf{12.77} &\bf{0.688} &\bf{0.0290} &\bf{0.9481} &0.0105\\
    \bottomrule
    \end{tabular}}
    }
    \caption{{\bf Ablation study on different label smoothing choice.} The first row is the model without label smoothing. Two-side: set 0 to 0.2 and 1 to 0.8; one-source: set 1 to 0.8; and 0ne-target: set 0 to 0.2. }
    \label{tab:label_smoothing}
 \end{table}

\section{Conclusion}
In this paper, we study the domain gap problem between clean product images and inpainted images in CGL-Dataset for generating poster layouts. To solve this problem, we propose to leverage the unsupervised domain adaptation technique and design a pixel-level discriminator. This design of discriminator can not only finely align image features of these two domains, but also avoid the Gaussian blurring step in the previous work~(CGL-GAN), which brings benefits to modeling the relationship between image details and layouts. Both quantitative and qualitative evaluations demonstrate that our method can achieve SOTA performance and generate high-quality image-aware graphic layouts for posters. In the future, we may investigate how to better interact with user constraints, e.g. categories and coordinates of elements, and enhance the layout generation diversity.

\section*{Acknowledgements}
This work is supported by Information Technology Center and State Key Lab of CAD\&CG, Zhejiang University.

{\small
\bibliographystyle{ieee_fullname}
\bibliography{CVPR2023}

\begin{thebibliography}{10}\itemsep=-1pt

\bibitem{DBLP:conf/cvpr/ArroyoPT21}
Diego~Mart{\'{\i}}n Arroyo, Janis Postels, and Federico Tombari.
\newblock Variational transformer networks for layout generation.
\newblock In {\em {CVPR}}, pages 13642--13652. Computer Vision Foundation /
  {IEEE}, 2021.

\bibitem{DBLP:conf/cvpr/BousmalisSDEK17}
Konstantinos Bousmalis, Nathan Silberman, David Dohan, Dumitru Erhan, and Dilip
  Krishnan.
\newblock Unsupervised pixel-level domain adaptation with generative
  adversarial networks.
\newblock In {\em {CVPR}}, pages 95--104. {IEEE} Computer Society, 2017.

\bibitem{DBLP:journals/tog/CaoCL12}
Ying Cao, Antoni~B. Chan, and Rynson W.~H. Lau.
\newblock Automatic stylistic manga layout.
\newblock {\em {ACM} Trans. Graph.}, 31(6):141:1--141:10, 2012.

\bibitem{DBLP:conf/mm/CaoMZLXGJ22}
Yunning Cao, Ye Ma, Min Zhou, Chuanbin Liu, Hongtao Xie, Tiezheng Ge, and
  Yuning Jiang.
\newblock Geometry aligned variational transformer for image-conditioned layout
  generation.
\newblock In {\em {ACM} Multimedia}, pages 1561--1571. {ACM}, 2022.

\bibitem{DBLP:conf/eccv/CarionMSUKZ20}
Nicolas Carion, Francisco Massa, Gabriel Synnaeve, Nicolas Usunier, Alexander
  Kirillov, and Sergey Zagoruyko.
\newblock End-to-end object detection with transformers.
\newblock In {\em {ECCV} {(1)}}, volume 12346 of {\em Lecture Notes in Computer
  Science}, pages 213--229. Springer, 2020.

\bibitem{DBLP:conf/iccv/CheferGW21}
Hila Chefer, Shir Gur, and Lior Wolf.
\newblock Generic attention-model explainability for interpreting bi-modal and
  encoder-decoder transformers.
\newblock In {\em {ICCV}}, pages 387--396. {IEEE}, 2021.

\bibitem{DBLP:journals/corr/abs-2010-03978}
Abolfazl Farahani, Sahar Voghoei, Khaled Rasheed, and Hamid~R. Arabnia.
\newblock A brief review of domain adaptation.
\newblock {\em CoRR}, abs/2010.03978, 2020.

\bibitem{DBLP:series/acvpr/GaninUAGLLML17}
Yaroslav Ganin, Evgeniya Ustinova, Hana Ajakan, Pascal Germain, Hugo
  Larochelle, Fran{\c{c}}ois Laviolette, Mario Marchand, and Victor~S.
  Lempitsky.
\newblock Domain-adversarial training of neural networks.
\newblock In Gabriela Csurka, editor, {\em Domain Adaptation in Computer Vision
  Applications}, Advances in Computer Vision and Pattern Recognition, pages
  189--209. Springer, 2017.

\bibitem{DBLP:journals/corr/Goodfellow17}
Ian~J. Goodfellow.
\newblock {NIPS} 2016 tutorial: Generative adversarial networks.
\newblock {\em CoRR}, abs/1701.00160, 2017.

\bibitem{DBLP:conf/nips/GoodfellowPMXWOCB14}
Ian~J. Goodfellow, Jean Pouget{-}Abadie, Mehdi Mirza, Bing Xu, David
  Warde{-}Farley, Sherjil Ozair, Aaron~C. Courville, and Yoshua Bengio.
\newblock Generative adversarial nets.
\newblock In {\em {NIPS}}, pages 2672--2680, 2014.

\bibitem{DBLP:conf/iccv/GuptaLA0MS21}
Kamal Gupta, Justin Lazarow, Alessandro Achille, Larry Davis, Vijay Mahadevan,
  and Abhinav Shrivastava.
\newblock Layouttransformer: Layout generation and completion with
  self-attention.
\newblock In {\em {ICCV}}, pages 984--994. {IEEE}, 2021.

\bibitem{DBLP:conf/cvpr/HeZRS16}
Kaiming He, Xiangyu Zhang, Shaoqing Ren, and Jian Sun.
\newblock Deep residual learning for image recognition.
\newblock In {\em {CVPR}}, pages 770--778. {IEEE} Computer Society, 2016.

\bibitem{DBLP:conf/cvpr/IsolaZZE17}
Phillip Isola, Jun{-}Yan Zhu, Tinghui Zhou, and Alexei~A. Efros.
\newblock Image-to-image translation with conditional adversarial networks.
\newblock In {\em {CVPR}}, pages 5967--5976. {IEEE} Computer Society, 2017.

\bibitem{DBLP:journals/tog/JacobsLSBS03}
Charles~E. Jacobs, Wilmot Li, Evan Schrier, David Bargeron, and David Salesin.
\newblock Adaptive grid-based document layout.
\newblock {\em {ACM} Trans. Graph.}, 22(3):838--847, 2003.

\bibitem{DBLP:conf/iccv/JyothiDHSM19}
Akash~Abdu Jyothi, Thibaut Durand, Jiawei He, Leonid Sigal, and Greg Mori.
\newblock Layoutvae: Stochastic scene layout generation from a label set.
\newblock In {\em {ICCV}}, pages 9894--9903. {IEEE}, 2019.

\bibitem{DBLP:conf/mm/KikuchiSOY21}
Kotaro Kikuchi, Edgar Simo{-}Serra, Mayu Otani, and Kota Yamaguchi.
\newblock Constrained graphic layout generation via latent optimization.
\newblock In {\em {ACM} Multimedia}, pages 88--96. {ACM}, 2021.

\bibitem{DBLP:conf/chi/KumarTAK11}
Ranjitha Kumar, Jerry~O. Talton, Salman Ahmad, and Scott~R. Klemmer.
\newblock Bricolage: example-based retargeting for web design.
\newblock In {\em {CHI}}, pages 2197--2206. {ACM}, 2011.

\bibitem{DBLP:conf/eccv/LeeJELG0Y20}
Hsin{-}Ying Lee, Lu Jiang, Irfan Essa, Phuong~B. Le, Haifeng Gong, Ming{-}Hsuan
  Yang, and Weilong Yang.
\newblock Neural design network: Graphic layout generation with constraints.
\newblock In {\em {ECCV} {(3)}}, volume 12348 of {\em Lecture Notes in Computer
  Science}, pages 491--506. Springer, 2020.

\bibitem{DBLP:conf/iclr/LiYHZX19}
Jianan Li, Jimei Yang, Aaron Hertzmann, Jianming Zhang, and Tingfa Xu.
\newblock Layoutgan: Generating graphic layouts with wireframe discriminators.
\newblock In {\em {ICLR} (Poster)}. OpenReview.net, 2019.

\bibitem{DBLP:journals/tvcg/LiY0LWX21}
Jianan Li, Jimei Yang, Jianming Zhang, Chang Liu, Christina Wang, and Tingfa
  Xu.
\newblock Attribute-conditioned layout {GAN} for automatic graphic design.
\newblock {\em {IEEE} Trans. Vis. Comput. Graph.}, 27(10):4039--4048, 2021.

\bibitem{DBLP:conf/cvpr/LinDGHHB17}
Tsung{-}Yi Lin, Piotr Doll{\'{a}}r, Ross~B. Girshick, Kaiming He, Bharath
  Hariharan, and Serge~J. Belongie.
\newblock Feature pyramid networks for object detection.
\newblock In {\em {CVPR}}, pages 936--944. {IEEE} Computer Society, 2017.

\bibitem{DBLP:conf/esann/MajumdarN18}
Debjeet Majumdar and Vinay~P. Namboodiri.
\newblock Unsupervised domain adaptation of deep object detectors.
\newblock In {\em {ESANN}}, 2018.

\bibitem{DBLP:journals/corr/abs-2205-12923}
Sushruth Nagesh, Shreyas Rajesh, Asfiya Baig, and Savitha Srinivasan.
\newblock Domain adaptation for object detection using {SE} adaptors and center
  loss.
\newblock {\em CoRR}, abs/2205.12923, 2022.

\bibitem{DBLP:journals/tvcg/ODonovanAH14}
Peter O'Donovan, Aseem Agarwala, and Aaron Hertzmann.
\newblock Learning layouts for single-pagegraphic designs.
\newblock {\em {IEEE} Trans. Vis. Comput. Graph.}, 20(8):1200--1213, 2014.

\bibitem{DBLP:conf/aaai/PeiCLW18}
Zhongyi Pei, Zhangjie Cao, Mingsheng Long, and Jianmin Wang.
\newblock Multi-adversarial domain adaptation.
\newblock In {\em {AAAI}}, pages 3934--3941. {AAAI} Press, 2018.

\bibitem{DBLP:conf/icml/RadfordKHRGASAM21}
Alec Radford, Jong~Wook Kim, Chris Hallacy, Aditya Ramesh, Gabriel Goh,
  Sandhini Agarwal, Girish Sastry, Amanda Askell, Pamela Mishkin, Jack Clark,
  Gretchen Krueger, and Ilya Sutskever.
\newblock Learning transferable visual models from natural language
  supervision.
\newblock In {\em {ICML}}, volume 139 of {\em Proceedings of Machine Learning
  Research}, pages 8748--8763. {PMLR}, 2021.

\bibitem{DBLP:journals/tip/RenLZH22}
Chuan{-}Xian Ren, Yong~Hui Liu, Xiwen Zhang, and Ke{-}Kun Huang.
\newblock Multi-source unsupervised domain adaptation via pseudo target domain.
\newblock {\em {IEEE} Trans. Image Process.}, 31:2122--2135, 2022.

\bibitem{DBLP:conf/wacv/SuvorovLMRASKGP22}
Roman Suvorov, Elizaveta Logacheva, Anton Mashikhin, Anastasia Remizova,
  Arsenii Ashukha, Aleksei Silvestrov, Naejin Kong, Harshith Goka, Kiwoong
  Park, and Victor Lempitsky.
\newblock Resolution-robust large mask inpainting with fourier convolutions.
\newblock In {\em {WACV}}, pages 3172--3182. {IEEE}, 2022.

\bibitem{DBLP:conf/cvpr/SzegedyVISW16}
Christian Szegedy, Vincent Vanhoucke, Sergey Ioffe, Jonathon Shlens, and
  Zbigniew Wojna.
\newblock Rethinking the inception architecture for computer vision.
\newblock In {\em {CVPR}}, pages 2818--2826. {IEEE} Computer Society, 2016.

\bibitem{DBLP:conf/nips/VaswaniSPUJGKP17}
Ashish Vaswani, Noam Shazeer, Niki Parmar, Jakob Uszkoreit, Llion Jones,
  Aidan~N. Gomez, Lukasz Kaiser, and Illia Polosukhin.
\newblock Attention is all you need.
\newblock In {\em {NIPS}}, pages 5998--6008, 2017.

\bibitem{DBLP:conf/cvpr/YangFYW21}
Cheng{-}Fu Yang, Wan{-}Cyuan Fan, Fu{-}En Yang, and Yu{-}Chiang~Frank Wang.
\newblock Layouttransformer: Scene layout generation with conceptual and
  spatial diversity.
\newblock In {\em {CVPR}}, pages 3732--3741. Computer Vision Foundation /
  {IEEE}, 2021.

\bibitem{DBLP:conf/cvpr/Zhang0TL22}
Jingyi Zhang, Jiaxing Huang, Zichen Tian, and Shijian Lu.
\newblock Spectral unsupervised domain adaptation for visual recognition.
\newblock In {\em {CVPR}}, pages 9819--9830. {IEEE}, 2022.

\bibitem{DBLP:journals/mta/ZhangD21}
Youshan Zhang and Brian~D. Davison.
\newblock Domain adaptation for object recognition using subspace sampling
  demons.
\newblock {\em Multim. Tools Appl.}, 80(15):23255--23274, 2021.

\bibitem{DBLP:journals/tog/ZhengQCL19}
Xinru Zheng, Xiaotian Qiao, Ying Cao, and Rynson W.~H. Lau.
\newblock Content-aware generative modeling of graphic design layouts.
\newblock {\em {ACM} Trans. Graph.}, 38(4):133:1--133:15, 2019.

\bibitem{DBLP:conf/ijcai/ZhouXMGJX22}
Min Zhou, Chenchen Xu, Ye Ma, Tiezheng Ge, Yuning Jiang, and Weiwei Xu.
\newblock Composition-aware graphic layout {GAN} for visual-textual
  presentation designs.
\newblock In {\em {IJCAI}}, pages 4995--5001. ijcai.org, 2022.

\end{thebibliography}
}

\end{document}